\documentclass[11pt,a4paper]{article}
\usepackage[hyperref]{naaclhlt2018}
\usepackage{tikz}
\usetikzlibrary{chains,fit,shapes,arrows,calc,shapes,decorations.pathreplacing}
\usepackage{subcaption}
\usepackage{times}
\usepackage{latexsym}
\usepackage{amssymb,amsmath,bm}
\usepackage{pgfplots}
\pgfplotsset{compat=1.4}
\usepackage{pgfplotstable}
\usepackage{arydshln}
\usepackage{url}

\DeclareCaptionFont{tenpt}{\fontsize{10pt}{11pt}\selectfont#1}
\DeclareMathOperator*{\argmax}{arg\,max}

\aclfinalcopy 

\title{Smaller Text Classifiers with
Discriminative Cluster Embeddings}

\author{Mingda Chen \qquad Kevin Gimpel \\
  Toyota Technological Institute at Chicago, Chicago, IL, 60637, USA \\
  {\tt \{mchen,kgimpel\}@ttic.edu} \\}

\date{}

\begin{document}
\maketitle
\begin{abstract}
Word embedding parameters often dominate overall model sizes in neural methods for natural language processing. We reduce deployed model sizes of text classifiers
by learning a hard word clustering in an end-to-end manner. We use the Gumbel-Softmax distribution to 
maximize over the latent clustering while minimizing the task loss. We propose variations that selectively assign additional parameters to words, which further improves accuracy while still remaining parameter-efficient. 
\end{abstract}

\section{Introduction}
Word embeddings~\citep{bengio2003neural} form the foundation of most neural methods for natural language processing (NLP). However, embeddings typically comprise a large fraction of the total parameters learned by a model, especially when large vocabularies and high dimensions are used. This can become problematic when seeking to deploy NLP systems on mobile devices where memory and computation time are limited. 

We address this issue by proposing alternative parameterizations for word embeddings in text classifiers. We introduce a latent variable for each word type that represents the (hard) cluster to which it belongs. An embedding is learned for each cluster. All parameters (including cluster assignment probabilities for each word and the cluster embeddings themselves) are learned jointly in an end-to-end manner. 

\begin{figure}[t]
    \centering
    \includegraphics[scale=1]{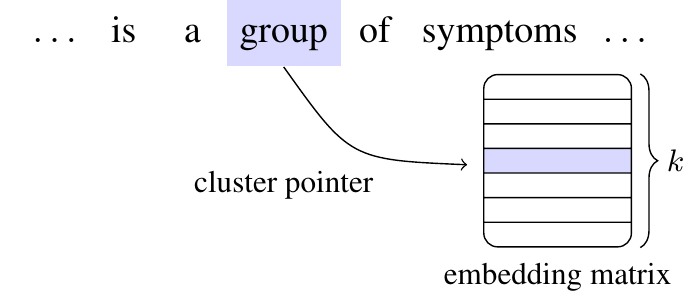}
    \label{fig1}
    \caption{Schematic of deployed cluster embedding model with $k$ clusters; cluster probabilities can be replaced by pointers at test time.\label{fig:schematic}}
\end{figure}

This idea is based on the conjecture that most words do not need their own unique embedding parameters, due both to the focused nature of particular text classification tasks and also due to the power law characteristics of word frequencies. For a particular task, many word embeddings would be essentially identical, so using clusters lets us avoid learning redundant embedding vectors, making parameter usage more efficient. For sentiment analysis, for example, the procedure can learn to place all sentiment-neutral words in a single cluster, and then learn distinct clusters for positive and negative words. 

During learning, we minimize log loss of the correct classification label while maximizing over the latent variables. To do so, we use the Gumbel-Softmax distribution~\citep{jang2016categorical,maddison2016concrete} as a continuous approximation to hard clustering. After training, we compute the argmax over cluster assignments for each word type and replace the cluster assignment probabilities with pointers to clusters; see Figure~\ref{fig:schematic}. This leads to a large reduction in model size at test time. 

We consider two variations of the above idea which introduce a small number of additional word-specific parameters. The best variation learns unique embeddings for only the most frequent words and uses hard clustering for the rest. We evaluate our methods on five text classification datasets, comparing them at several model size budgets. 
Our results demonstrate that clustering can maintain or improve performance while offering extremely small deployed models. 

\section{Related Work}

Several methods have been proposed for reducing the memory requirements of models that use word embeddings. One is based on quantization \cite{botha2017natural,han2015deep},
which changes the way parameters are stored. 
In particular, it seeks to find shared weights among embedding vectors and only keeps scale factors for each word.

Another family of methods uses hashing functions to replace dictionaries~\cite{svenstrup2017hash,joulin2016fasttext}. This can save storage space,
but still requires the model to have roughly the same size embedding matrix.
Network pruning has also been used to compress neural networks.  \citet{han2015learning} pruned weights iteratively by removing weights below a threshold and then retraining the network.

Our work is also related to prior work using hard word clustering for NLP tasks~\citep{botha2017natural,brown1992class}. The primary difference is that we cluster words to minimize the task loss rather than doing so beforehand.

Recently, \citet{shu2017compressing} also found clustering helpful for compressing the word embedding matrix for NLP tasks. Their method follows the intuition of product quantization \cite{jegou2011product,joulin2016fasttext}. Our methods differ from theirs in two ways. First, our methods are trained end-to-end instead of relying on pretrained word embeddings. Since our embeddings are trained for each task, we can use much smaller embedding dimensionalities, which saves a lot of parameters. Second, our method is faster at test time because it does not use multiple code books.

\section{Embedding Parameterizations}

Our text classifiers use long short-term memory (LSTM; \citealp{hochreiter1997long}) networks to embed sequences of word embeddings and then use the final hidden state as input to a softmax layer to generate label predictions. The standard cross entropy loss is used for training. We use this same architecture throughout and vary the method of parameterizing the word embedding module among the four options listed below. 

\paragraph{Standard Embeddings (SE).} This is the standard setting in which each word type in the vocabulary has a unique embedding. Given a vocabulary $V=\{w_1,w_2,\cdots,w_v\}$ and embedding dimensionality $m$, this yields $vm$ word embedding parameters. In our experiments, we limit the vocabulary to various sizes $v$, always keeping the most frequent $v$ words and replacing the rest with an unknown word symbol. 

\paragraph{Cluster Embeddings (CE).} 
We next propose a method in which each word is placed into a single cluster (``hard clustering'') and we learn a unique embedding vector for each cluster. We refer to this setting as using cluster embeddings (CE). 
We denote the embedding matrix $W\in\mathbb{R}^{k\times m}$ where $k$ is the number of clusters and $m$ is again the embedding dimensionality. 
Each word $w_i$ now has a vector of parameters $\vec{a_i}=(a_{i1},\cdots,a_{ik})$ which are interpreted as cluster probabilities. So this method requires learning $vk$ cluster probabilities in addition to the $km$ parameters for the cluster embeddings themselves. 

We treat the cluster membership of each word $w_i$ as a latent variable $h_i$ during training. All parameters are optimized jointly to minimize cross entropy while maximizing over the latent variables. This poses difficulty in practice due to the discrete nature of the clustering. That is, maximizing over the latent variables involves non-differentiable argmax operations:
\begin{equation}
h_i = \argmax_{1\leq j \leq k} \, a_{ij}
\nonumber
\end{equation}
\noindent To tackle this problem, we use the recently proposed Gumbel-Softmax to approximate the clustering decision during training. Gumbel-Softmax is a temperature-modulated continuous relaxation for the categorical distribution. When the temperature approaches $0$, samples from Gumbel-Softmax will become identical to those from the categorical distribution. During training, we have a sample $\vec{t_i}=(t_{i1},\cdots,t_{ik})$ for every instance of a word $w_i$. The vector $\vec{t_i}$ is a non-sparse approximation to the one-hot vector indicated by the latent variable value $h_i$. It is parameterized as:

\begin{equation}
    t_{ij}=\frac{\exp{(( a_{ij} + g_j)/\tau)}}{\sum_{l=1}^k \exp{((a_{il} + g_l)}/\tau)}\nonumber
\end{equation}
\noindent where the $g_j$ are samples from a Gumbel(0,1) distribution and $\tau$ is the temperature. The embedding vector $\vec{e_i}$ for word $w_i$ is calculated by $\vec{e_i}=W^\top \vec{t_i}$.

Even when merely using this method in a soft clustering setting, it can save parameters when $vk + km < vm$. But with hard clustering, we can reduce this further by assuming we will again maximize over latent variables at test time. In this case, the cluster for word $w_i$ at test time is
\begin{equation}
  \vec{t_i}=\text{one\_hot}\left(\argmax_{1\leq j \leq k} \, a_{ij}\right)\nonumber
\end{equation}
\noindent where the function $\text{one\_hot}$ returns a one-hot vector of length $k$ with a 1 in the index given by its argument. These argmax operations can be precomputed for all words in $V$, permitting us to discard the $vk$ cluster probabilities and instead just store a cluster pointer for each word, each of which will only take $O(\log_2 k)$ space. 

\paragraph{Cluster Adjustment Embeddings (CAE).}  
While the cluster embedding model can lead to large savings in parameters, it loses the ability to model subtle distinctions among words, especially as $k$ decreases. We propose a modification (cluster adjustment embeddings; CAE) that represents a word by concatenating its cluster embedding with a short unique vector for the word.  If we think of cluster embeddings as centroids for each cluster, this model provides a way to adjust or correct the cluster embedding for each word, while still leveraging parameter sharing via the cluster embeddings. For all CAE experiments below, we use a 1-dimensional vector (i.e., a scalar) as the unique vector for each word that gets concatenated to the cluster embedding. 

\paragraph{Mixture Embeddings (ME).} 
Finally, we consider a variation (mixture embeddings; ME) in which the most frequent $u$ words use unique embeddings and the remaining words use cluster embeddings. The words with unique embeddings are selected based on word frequency in the training data, with the intuition that frequent words are potentially useful for the task and contain enough instances to learn unique embedding parameters.

\section{Experimental setup}
\begin{table}[t]
\setlength{\tabcolsep}{3pt}
\centering
\small
\begin{tabular}{lcccc}
  Dataset & \# Classes & Train & Dev. & Test \\
  \hline
  AG News & 4 & 115,000 & 5,000 & 7,600 \\
  DBpedia & 14 & 560,000 & 5,000 & 70,000 \\
  Yelp Review Polarity & 2 & 555,000 & 5,000 & 38,000 \\
  Yelp Review Full & 5 & 645,000 & 5,000 & 50,000 \\
  IMDB Movie Reviews & 2 & 23,000 & 2,000 & 25,000 \\
\end{tabular}
\caption{\label{data_summary} Dataset statistics.}
\end{table}

We evaluate our embedding models on five text classification datasets: AG News, DBpedia, Yelp Review Polarity, Yelp Review Full~\cite{zhang2015character}, and the IMDB movie review dataset \cite{maas2011learning}. We randomly sample 5,000 instances from the 
training set to use as development data for all datasets except for IMDB, where we sample 2,000. 
Table~\ref{data_summary} shows dataset statistics.
For IMDB, to make our results comparable to \citet{shu2017compressing}, we follow their experimental setup: We tokenize and lowercase the IMDB data using NLTK 
and truncate each review to be at most 400 words. For the other datasets, we lowercase and tokenize the sentences using regular expressions based on \citet{kim2014convolutional}. 

For optimization, we use Adam \cite{kingma2014adam} with learning rate 0.001. Embedding matrices are randomly initialized for all models. 
To reduce the hyperparameter search space, the LSTM hidden vector size is set to 50 for all experiments and the Gumbel-Softmax temperature is fixed to 0.9. 
When a single result is reported, all other hyperparameters (vocabulary size $v$, embedding dimension $m$, number of clusters $k$, and number of unique vectors $u$) are tuned based on the  development sets. 
Our code is implemented in TensorFlow \cite{tensorflow2015-whitepaper} and is available at \url{https://github.com/mingdachen/word-cluster-embedding}.

\section{Results}

\begin{figure}[t]
    \centering
    \begin{subfigure}{.25\textwidth}
    \includegraphics[width=1.0\linewidth]{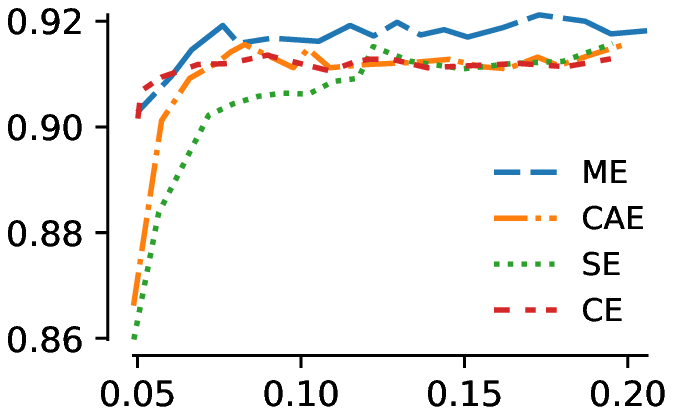}
     \caption{AG News}
     \label{fig2:ag}
    \end{subfigure}%
    \begin{subfigure}{.25\textwidth}
    \includegraphics[width=1.0\linewidth]{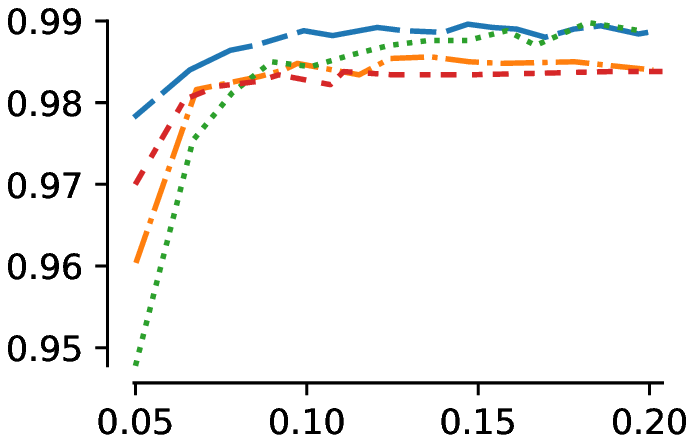}
     \caption{DBpedia}
     \label{fig2:dbp}
    \end{subfigure}

    \begin{subfigure}{.25\textwidth}
    \includegraphics[width=1.0\linewidth]{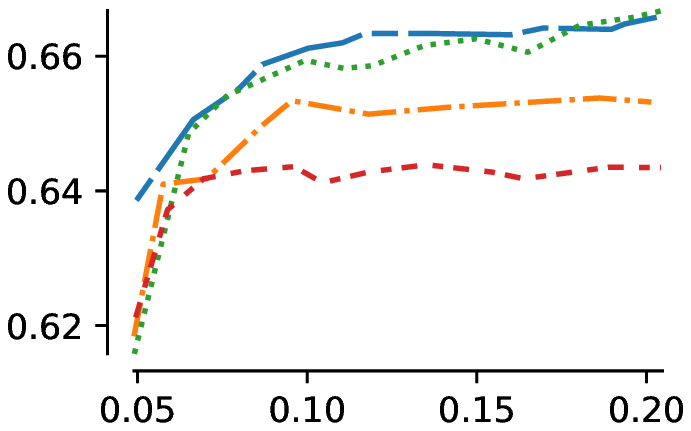}
     \caption{Yelp Full}
     \label{fig2:y1}
    \end{subfigure}%
    \begin{subfigure}{.25\textwidth}
    \includegraphics[width=1.0\linewidth]{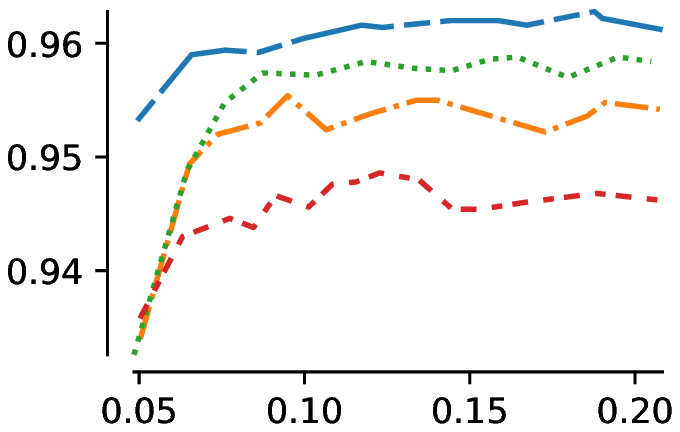}
     \caption{Yelp Polarity}
     \label{fig2:y2}
    \end{subfigure}%
    \caption{Development accuracy vs model size (MB) on four datasets. ME consistently outperforms other models under various size budgets.}
    \label{fig2}
\end{figure}

\begin{table}[t]
\setlength{\tabcolsep}{3pt}
\centering
\small
\begin{tabular}{l|cc|cc|cc|cc}
        & \multicolumn{2}{c|}{AG News} &  \multicolumn{2}{c|}{DBPedia} & \multicolumn{2}{c|}{Yelp Full} & \multicolumn{2}{c}{Yelp Polarity} \\
\hline
size  & 0.05 & 0.1 &                     0.05    &  0.1 &                        0.05 & 0.1 &                    0.05 & 0.1 \\ 
\hline
SE  & 84.8 & 90.4 &    95.3 & 98.1 &   59.2 & 62.6    & 93.4 & 95.5 \\
CE  & 89.2 & 90.7 &    96.9 & 97.9 &   60.3 & 61.0    & 93.9 & 94.4 \\
CAE & 86.3 & 90.7 &    96.1 & 98.1 &   61.2 & 62.3    & 93.7 & 95.3 \\
ME  & \textbf{90.3} & \textbf{91.5} &    \textbf{97.5} & \textbf{98.3} &   \textbf{61.4} & \textbf{63.4}    & \textbf{95.2} & \textbf{95.8} \\
\end{tabular}
\caption{Test results. Model sizes are in MB.} 
\label{all-res}
\end{table}

When evaluating our models, we are concerned with both accuracy and model size. We vary hyperparameters to obtain a range of model sizes for each embedding parameterization, then train models for each set of hyperparameter values. In Figure~\ref{fig2}, we plot development accuracies across the range of model sizes on four datasets. Model sizes are calculated using the formula given in the appendix. 

When the model size is extremely small (e.g., less than 0.1 MB in AG News), our cluster models outperform the standard parameterization (SE). As model size increases, the standard model becomes better and better, though it does not outperform ME. While CE is weak on the Yelp tasks, which could be attributed to the difficulty of 5-way sentiment classification, we see clear improvements by adding small amounts of word-specific information via CAE and ME. ME consistently performs well compared to the others across model sizes. 

We report test results in Table~\ref{all-res}. 
The test results are reported based on model performance on the development set for different model sizes. The models are consistent between development and test, as our cluster models with max size 0.05MB outperform SE across datasets, with ME having the highest accuracies.

On IMDB, we compare our methods to compositional coding~\cite{shu2017compressing}. This method learns an efficient coding of word embeddings via the summation of embeddings from multiple clusterings. The clusterings and cluster embeddings are learned offline to reconstruct pretrained GloVe vectors~\citep{pennington2014glove}. We recalculated the embedding sizes from \citet{shu2017compressing} using our formula (in the appendix). We also reimplemented their compositional coding as another embedding model and trained it in an end-to-end fashion. We use the best model configuration from \citet{shu2017compressing} and do grid search for the embedding dimension. As for the vocabulary size $v$, we find models perform better with small values, and thus we fix it to $v=3000$. 

\begin{table}[t]
\setlength{\tabcolsep}{1pt}
\centering
\small
\begin{tabular}{lccc}
        & \multicolumn{1}{c}{embedding} & model & acc. \\
        & \multicolumn{1}{c}{size (MB)}    & size (MB) &  (\%)       \\
\hline
GloVe baseline        & 85.947          & -          & 87.18          \\
8 $\times$ 8 coding   & 0.288           & -          & 82.84          \\
16 $\times$ 32 coding & 1.302           & -          & 87.37          \\
64 $\times$ 8 coding  & 2.305           & -          & 88.15          \\ \hdashline
64 $\times$ 8 coding ($m=90$)  & 0.245           & 0.353    & 83.43          \\
SE ($|V|=3000$, $m=8$)             & 0.092           & 0.137    & 86.84          \\
CE ($k=50$, $m=5$)               & \textbf{0.004} & 0.046    & 85.58          \\
CAE ($k=50$, $m=5$)              & 0.016           & 0.058    & 86.94          \\
ME ($k=50$, $u=300$, $m=5$)           & 0.009           & 0.051    & \textbf{88.22}
\end{tabular}
\caption{IMDB test results. The four rows above the dashed line are from \citet{shu2017compressing}; our results are below it.} 
\label{imdb-res}
\end{table}

Results are shown in Table~\ref{imdb-res}. Compared with compositional coding, our models perform much better with a much smaller set of embedding parameters even when we use a smaller number of cluster embeddings (e.g., compare 8 $\times$ 8 coding to CE; both use a comparable number of cluster embedding vectors, while CE works better). ME (with $k=50$ clusters and unique embeddings for the $u=300$ most frequent words) outperforms all other models while retaining a small model size. CAE performs better than SE, but uses more parameters than CE. We find a better trade-off with ME, which only adds parameters to the most frequent words. 

\section{Discussion}

\subsection{Cluster Analysis}
Table~\ref{tb3} shows clusters learned via CE on the AG News dataset. Cluster 1 appears to contain words that are related to quantities such as times and numbers while cluster 2 mostly contains prepositions and other function words. 
The connection to the AG News labels (World, Sports, Business, and Sci/Tech) is more clear in the subsequent clusters. Clusters 3 and 8 are related to World, cluster 4 may relate to World or Sports,  clusters 5 and 6 are related to Sci/Tech, and cluster 7 is related to Sports.
\begin{table}[t]
\setlength{\tabcolsep}{5pt}
\centering
\small
\begin{tabular}{l|l}
   & \bf words \\
  \hline
  1 & million week third percent which 000 ago reports once \\
  2 & are after from has another down home than but end \\
  3 & official security china international country court city  \\
  4 & heavyweights operational coordinated healing rewarded
  \\
  5 & com internet technology ibm google research windows \\
  6 & market quarter sales deals bid growth trade economic \\
  7 & championship yankees defense player contract football \\
  8 & troops press attack forces peace iran led army killing \\
\end{tabular}
\caption{\label{tb3} Word clusters learned using CE model on AG News. Each row is a different cluster.}
\end{table}

\subsection{Impact of Hyperparameters}

\begin{figure}[t]
    \centering
    \begin{subfigure}{.25\textwidth}
    \includegraphics[width=1.0\linewidth]{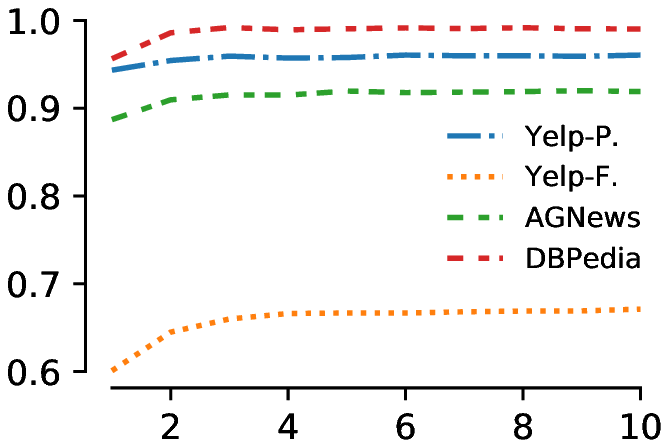}
    \caption{SE embedding dim. ($m$)}
    \label{fig3:a}
    \end{subfigure}%
    \begin{subfigure}{.25\textwidth}
    \includegraphics[width=1.0\linewidth]{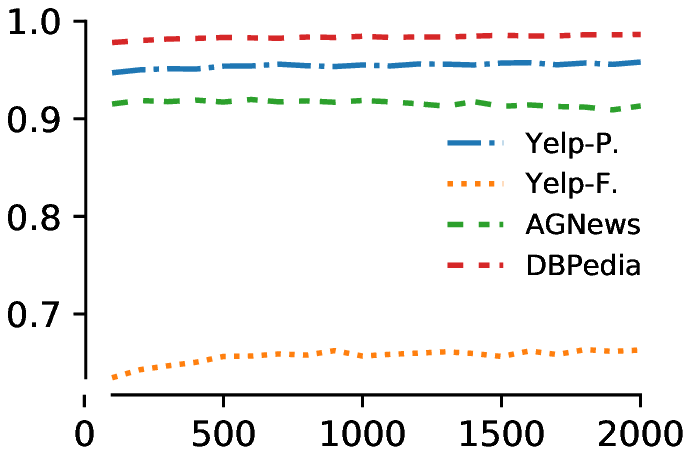}
    \caption{ME \# unique emb. ($u$)}
    \label{fig3:b}
    \end{subfigure}

    \begin{subfigure}{.25\textwidth}
    \includegraphics[width=1.0\linewidth]{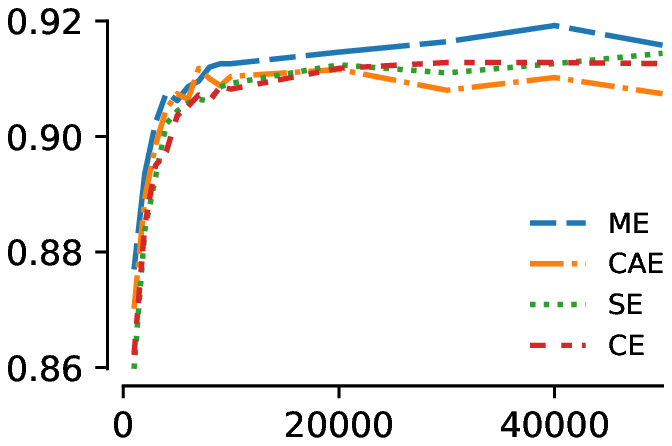}
    \caption{Vocab. size (AG News)}
    \label{fig3:c}
    \end{subfigure}%
    \begin{subfigure}{.25\textwidth}
    \includegraphics[width=1.0\linewidth]{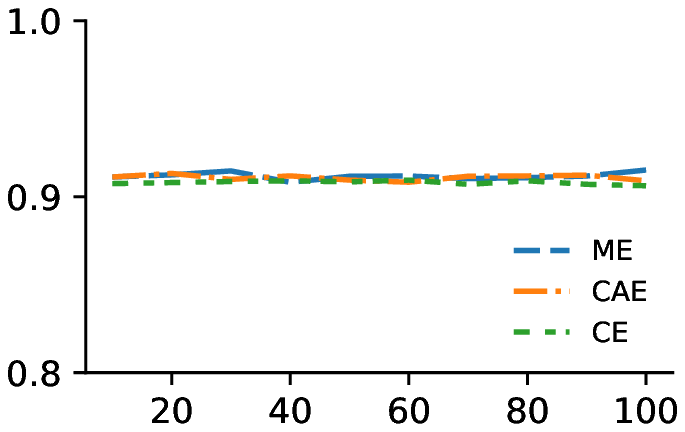}
    \caption{\# clusters (AG News)}
    \label{fig3:d}
    \end{subfigure}%
    \caption{Dev.~accuracy vs.~hyperparameters.}
    \label{fig3}
\end{figure}

\noindent Figure \ref{fig3} shows the relationship between accuracy and several hyperparameters. Figure \ref{fig3:a} shows the effect of embedding dimension on SE models. One-dimensional embeddings work reasonably well, but the largest accuracy gains occur when increasing dimensionality from 1 to 2. Consider the LSTM gating functions, which consist of a nonlinearity applied to $U_1 \mathbf{x} + U_2 \mathbf{h}$, where $\mathbf{x}$ is a word embedding, $\mathbf{h}$ is a hidden state, and $U_1,U_2$ are parameters. We can think of these functions as doing affine transformations on the hidden state. So, in the one-dimensional case, the transformations that a word vector can do are restricted to translation. However, when word vectors have more than two dimensions, they can do almost any affine transformation. To further investigate this, we experimented with simple recurrent neural networks (RNNs) with very small word embedding dimensionalities in the appendix.

Figure~\ref{fig3:b} shows that for most datasets, increasing the number of unique embedding vectors ($u$) in ME helps for the Yelp datasets, especially early on, but $u=500$ appears sufficient to capture most of the accuracy. Since similar trends are observed across different datasets, we only plot results for AG News in the final two plots. In Figure \ref{fig3:c}, there is a clear boundary after which vocabulary size has minimal effect on accuracy. In Figure \ref{fig3:d}, we observe that the number of clusters does not have much impact.

The main differences among CE, CAE, and ME are the ways they balance precision of embedding vectors and overall model size. CE forgets word identities and uses common parameters for all words in a cluster. Therefore it is expected that it should perform best only when all models are restricted to be extremely small. CAE adds parameters evenly across all words in the vocabulary while ME focuses its additional parameters on the most frequent words. Our results show that devoting parameters to the most frequent words achieves the best balance and consistently strong results. The most frequent words in the training set are likely to be those most closely related to the task. Higher frequency also means more training data for the word's embedding parameters.

\subsection{Impact of Training Data Size}
\label{app:datasize}

\begin{figure}[t]
    \centering
    \includegraphics[ scale=0.56]{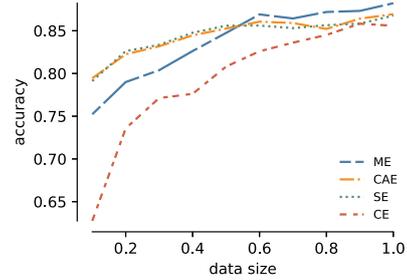}
    \caption{Varying the fraction of training data used on the IMDB task.}
    \label{fig5}
\end{figure}

Figure~\ref{fig5} shows test accuracies when varying the size of the training set for the IMDB task. The clustering models need relatively large amounts of training data, because they actually may have more parameters to learn during training due to the cluster membership probabilities for each word. We suspect this is why ME underperforms SE and CAE with small training sets. Even though ME permits very small deployed models, it still requires a substantial training set to learn its cluster membership probabilities.

\section{Conclusions and Future Work}

We proposed word embedding parameterizations that dramatically reduce the number of parameters at test time while achieving comparable or better performance. Our methods are applicable to other neural methods that use word embeddings or any kind of parameter lookup data structure. Future work will incorporate pretrained word embeddings into these cluster parameterizations and apply them to additional tasks.


\section*{Acknowledgments}
We thank Qingming Tang for helpful discussions and the anonymous reviewers for their comments. 

\bibliography{naaclhlt2018}
\bibliographystyle{acl_natbib}

\appendix

\section{Model Size Calculation}
\label{app:eq1}

Let the model have vocabulary size $v$, $k$ embedding vectors, embedding dimension $m$, and $o$ other parameters. To compute model sizes, we assume each cluster pointer is stored using $\lceil\log_2{k}\rceil$ bits and that other parameters are stored using 32 bits. For the CE model, for example, model size can be calculated based on the following formula:
\begin{equation}
    v*\lceil\log_2{k}\rceil + k * m * 32 + o * 32\nonumber
\end{equation}

\section{Impact of Word Embedding Dimension}
\label{app:emb}

\begin{figure}[htp]
    \centering
    \begin{subfigure}{.25\textwidth}
    \includegraphics[width=1.0\linewidth]{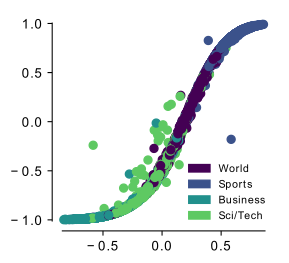}
     \caption{1-dim.~word embeddings}
     \label{fig4:1}
    \end{subfigure}%
    \begin{subfigure}{.25\textwidth}
    \includegraphics[width=1.0\linewidth]{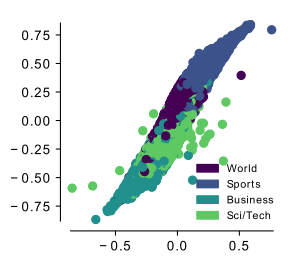}
     \caption{2-dim.~word embeddings}
     \label{fig4:2}
    \end{subfigure}
    
    \begin{subfigure}{.25\textwidth}
    \includegraphics[width=1.0\linewidth]{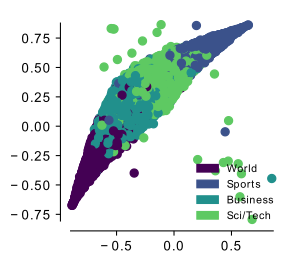}
     \caption{3-dim.~word embeddings}
     \label{fig4:3}
    \end{subfigure}%
    \begin{subfigure}{.25\textwidth}
    \includegraphics[width=1.0\linewidth]{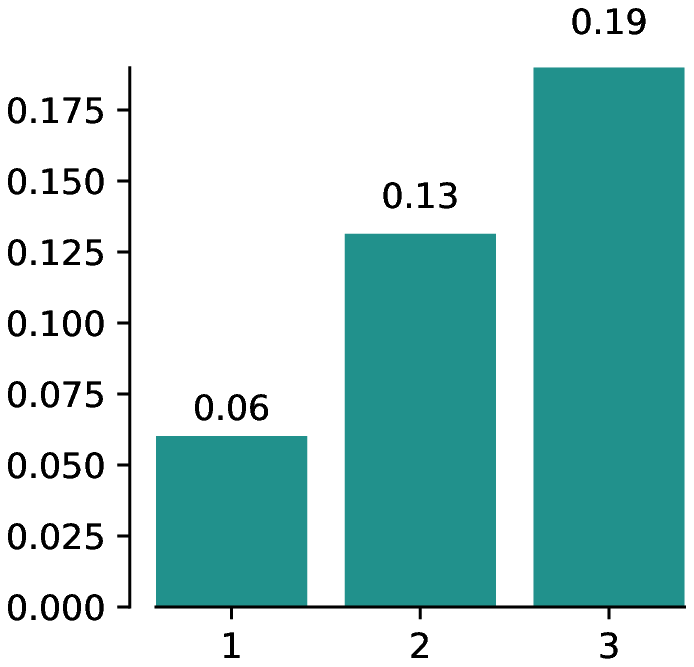}
     \caption{Area ratio vs.~dimension}
     \label{fig4:4}
    \end{subfigure}%
    \caption{Plots of 2-dimensional RNN hidden states when varying word embedding dimensionality.}
    \label{fig4}
\end{figure}

\begin{table}[t]
\centering
\begin{tabular}{lccc}
Embedding Dimension & 1    & 2    & 3    \\ \hline
Accuracy            & 0.82 & 0.78 & 0.84
\end{tabular}
\caption{IMDB test results for RNN with different embedding dimensions.}
\label{rnn-res}
\end{table}

In order to look into the impact of word embedding dimension, we run experiments on AG News using an RNN with 2-dimensional hidden states instead of an LSTM. Figure \ref{fig4} plots the final hidden states of the RNN with various embedding dimensions.

When embeddings have one dimension (Figure~\ref{fig4:1}), most of the hidden states roughly lie on a line, which is expected considering the limited transformation a scalar can do. As the dimension increases, the hidden states become more spread out. To evaluate this phenomena quantitatively, we calculate the area ratio that hidden states have covered, which is shown in Figure~\ref{fig4:4}. The area ratio increases monotonically with increasing embedding dimension. We also report the corresponding test accuracies in Table~\ref{rnn-res}. The classification accuracy does not necessarily improve from larger usage of space. The reason for this could be the vanishing gradient problem in simple RNNs.

\end{document}